%% file: main.tex
\documentclass[letterpaper]{article} 
\pdfoutput=1
\usepackage{aaai23}  
\usepackage{times}  
\usepackage{helvet}  
\usepackage{courier}  
\usepackage[hyphens]{url}  
\usepackage{graphicx} 
\usepackage{natbib}  
\usepackage{caption} 
\usepackage{algorithm}
\usepackage{algorithmic}
\usepackage{verbatim}
\usepackage{xcolor}
\usepackage[T1]{fontenc}

\usepackage{subcaption}

\newcommand{\secref}[1]{Section~\ref{sec:#1}}
\newcommand{\seclabel}[1]{\label{sec:#1}}

\newcommand{\eg}{\textit{e.g.,}}

\usepackage{newfloat}
\usepackage{listings}
\DeclareCaptionStyle{ruled}{labelfont=normalfont,labelsep=colon,strut=off} 
\floatstyle{ruled}
\newfloat{listing}{tb}{lst}{}
\floatname{listing}{Listing}
\title{Responsible Robotics: A Socio-Ethical Addition To Robotics Courses}

\author{Joshua Vekhter, Joydeep Biswas}
\affiliations{
    The University of Texas at Austin\\
    2317 Speedway, Stop D9500, Austin, TX, USA 78712\\
    josh@cs.utexas.edu,joydeepb@cs.utexas.edu

}

\begin{document}

\maketitle

\begin{abstract}
  We are witnessing a rapid increase in real-world autonomous robotic deployments in environments ranging from indoor homes and commercial establishments to large-scale urban areas, with applications ranging from domestic assistance to urban last-mile delivery. The developers of these robots inevitably have to make impactful design decisions to ensure commercial viability, but such decisions have serious real-world consequences. Unfortunately, it is not uncommon for such projects to face intense bouts of social backlash, which can be attributed to a wide variety of causes, ranging from inappropriate technical design choices to transgressions of social norms and lack of community engagement.

  To better prepare students for the rigors of developing and deploying real-world robotics systems, we developed a \emph{Responsible Robotics} teaching module, intended to be included in upper-division and graduate-level robotics courses. Our module is structured as a role-playing exercise that aims to equip students with a framework for navigating the conflicting goals of human actors which govern robots in the field. We report on instructor reflections and anonymous survey responses from offering our responsible robotics module in graduate-level and upper-division undergraduate robotics courses at UT Austin. The responses indicate that students gained a deeper understanding of the socio-technical factors of real-world robotics deployments than they might have using self-study methods, and the students proactively suggested that such modules should be more broadly included in CS courses.

\end{abstract}

\section{Introduction}
As AI research continues to advance state of the art results at breakneck speeds, many researchers have begun advocating for the necessity of including ethics modules in technical courses~\cite{role_of_ethics}.
Much as medical professionals are required to take courses on medical ethics before they are allowed to see patients, the software engineers and designers who implement AI systems wield great power to shape society and we all stand to benefit when their decision making is grounded in sound ethical principles, such as those codified into the ACM code of ethics~\cite{ACM_ethics_code}.

 In and out of the classroom, it has become increasingly common for AI
 researchers and practitioners to ask questions regarding bias, fairness, and
 privacy when designing and deploying AI powered services.  While this is a
 commendable first step, these three factors alone are insufficient in capturing
 the ethical challenges involved in deploying robots in real-world human
 environments\footnote{Making algorithmic and hardware design choices based on
 simplified models of target deployment environments is inevitable and in fact a
 necessary part of an effective design process.  But these choices then place
 constraints on the environments where robots can be safely deployed.  Even when
 a responsible roboticist is acting in persuit of commercial viability, they
 should strive to consider the human impacts of their design decisions.
 }. The hardware aspect of robotics raises many issues and failure modes not present in software products.  We identify \emph{safety}, \emph{local community sentiment}, \emph{disparate impact}, and \emph{labor relations} as four key additional ethical factors crucial to robotics.





We believe that there is a clear demand for specialized university curricula that aims to discuss the social impacts of technical decision making, in a setting which is \emph{embedded} within the context of a \emph{technical} robotics curriculum.

We contribute a \emph{Responsible Robotics} module designed to fit into existing robotics courses.  This module begins to address the aforementioned ethical considerations, their relationship to technical design decisions, and the roles and responsibilities of roboticists today. Our Responsible Robotics module consists of three parts:
\begin{enumerate}
    \item A \emph{Required Reading} section to familiarize students with the issues surrounding ethical robotics deployments
    \item A \emph{Written Essay} section to encourage students to think about the interactions between the issues and technical design decisions
    \item A series of in-class \emph{Role-Playing} scenarios to reason through the issues in practice, and their ramifications on society at large.
\end{enumerate}
 We document our curricular design, instructor experience, and student feedback here in the hope that other educators will use it to inform the inclusion of socio-ethical factors in their own robotics courses.

 This paper is organized as follows: \secref{context} describes the type of technical course for which the responsible robotics module is designed; \secref{related_work} relates our contributions to previously developed related courses and modules; \secref{module} describes the curricular objectives, design decisions, and the structure of the module; \secref{activities} describes the in-class activities in detail; and \secref{evaluation} presents evaluations from anonymous surveys and instructor reflections.

\section{Course Context}
\seclabel{context}
The Responsible Robotics module was explicitly designed to be embedded within a \emph{technical} robotics course.  We contend that not only should technical courses include discussions on socio-ethical factors, but also that serious discussion of complex socio-ethical factors of robotics \emph{requires} understanding concrete implementation level details that limit the performance of algorithms and hardware.

There are numerous technical courses that train students to understand the low
level impact of algorithmic changes on system performance. For example, in robotics, it
is common to cover particle filters~\cite{fox2001particle}, where students might
learn that that adding more particles means higher robustness and accuracy at
the cost of increased compute.  However, rarely are students asked to consider
socio-ethical impacts of such algorithmic choices.  For example, a last-mile
delivery robot relying on a particle filter with too few particles for reduced
computational cost may mistake a
neighbor's driveway for the intended recipient's driveway, and deliver a
potentially sensitive package to the wrong address. The consequence of such a
technical failure is significantly more egregious than just a failure to
complete a task  --- it compromises the privacy of potential customers.

Outside a technical robotics course, it may be instructive to list socio-ethical factors related to deploying robots such as disparate impact and safety, but to gain a deep understanding of \emph{how} such factors manifest in real robot deployments, it is necessary to understand what algorithms are used, their assumptions and limitations, and the conscious technical design decisions that go into deploying such algorithms on real robots.
For example, a learned visual feature extractor~\cite{detone2018superpoint} used for visual simultaneous localization and mapping~\cite{deng2019comparative} may be most effective in dense urban settings with ample human-made structures, but less effective in rural environments with fewer geometric features. It may be tempting to use such a technical justification in order to limit robotic deployments to downtown areas alone, but this now means that the robot on-which it is deployed is unsuitable for deployment in rural areas.  Further, not all urban environments are created equal, \eg{} some are highly unstructured or poorly maintained, while others are highly accessible~\cite{knight_found}.  Here again hardware design choices may limit deployments in ways that creates \emph{disparate impacts}.

We deployed our module in two courses, one aimed at graduate students, the other at upper division undergrads.  Both courses were intended to serve as an overview of robotics, with an emphasis on perception and planning. We studied algorithms and data structures related to these topics, including state estimation, mobile robot localization, simultaneous localization and mapping, planning, and motion control. Students gain hands-on experience in implementing and extending such algorithms on a fleet of scale 1/10 autonomous cars~\cite{ut_automata}.
We devoted one assignment (out of five total) and a week in the semester to the responsible robotics module. The objective of including this module in the curriculum was to get students to think about the broader implications of deploying robots in society, and the implications of technical decisions. To this end, we developed an ``Inquiry Based Learning'' assignment that provided students with an opportunity to form their own conclusions and to grapple with the open-ended nature of the questions raised in class while behaving respectfully~\cite{rules_of_engage}.


\section{Related Work}
 \seclabel{related_work}

We review previous initiatives for incorporating ethics into computing, AI and ML, and most recently in robotics; and summarize how our responsible robotics module relates to the previous work.

\paragraph{Early Years of Ethics Education in Computer Science}
The ACM first introduced a codified code of ethical conduct in 1972, which was then updated in 1992 and again in 2018 to reflect the changing reality of computing.  In 1991, the ACM released a curriculum report which recommended that students receive $\sim{}11$ hours of instruction on ethical and social issues as part of an undergraduate education~\cite{ACM_1991_curicula_report}.  However practitioners at the time felt that the guidance in this report was too vague, and in 1994 a panel of 25 experts convened to develop concrete ethics modules for educators as part of the ImpactCS project~\cite{integrating_ethics_into_cs_1997}, and now recommended that students receive 50-60 hours of ethics instruction.  A practical teaching guide from the same year advocates for the effectiveness of using role-playing as a teaching tool~\cite{getting_started_ethics_1997}.

\paragraph{Responsible Computing Builds Upon Sound Ethics}
Skip forward 25 years, we see that despite considerable effort, ethics education has not been broadly integrated into CS curricula for a variety of reasons. While this is only a small snapshot of current on-going efforts, we note that a number of educators, including a group at Harvard~\cite{harvard2019} have begun advocating for integrating ethics modules into every CS course. In 2022, a large report from the National Academy of Sciences~\cite{NAP26507} was published, advocating for greater consideration of socio-ethical concerns across the entire field of computing, from training to research to practice, and suggested considering ethical impact when making future funding decisions.

In a parallel thread, between 2018-2021, Mozilla also funded a large grant for developing \emph{Responsible Computer Science} curricula~\cite{MOZilla_resp_cs}.  One project to come out of this work was the ``Embedding Ethics in CS Classes Through Role Play'' project from Georgia Tech~\cite{gatech_resp_cs}, which we used as a source of inspiration when developing the module presented here.  Another role-playing approach to CS education which informed our work is a seminar structure in the computer graphics community~\cite{Jacobson2021RolePlayingPS}.  Yet another example is a case-study based approach to teaching AI ethics~\cite{Kuipers_ai_ethics}.





\paragraph{Ethical Robotics}

When considering robotics in particular, we note that a large fraction of the ``Main Debates'' listed in the current entry for ``\emph{Ethics of Artificial Intelligence and Robotics}'' in the Stanford Encyclopedia of Philosophy~\cite{stanford-ethics-ai} involve robotics in some way.  Once an autonomous agent begins to directly control atoms, we are forced to attempt to codify otherwise ephemeral scenarios in an attempt to make robot agents safe and trustworthy~\cite{how_to_trust_robot}, in an attempt to ensure that it's behavior is \emph{well aligned} to the ethical principles which govern the society in which it is deployed~\cite{DBLP:conf/aaaiss/Kuipers16}. The ethical issues that appear when dealing with autonomous agents are fundamentally different than those which arise in purely software based systems and both merit discussion in a complete degree program.

\paragraph{Robots in Public Spaces}
One of the most challenging unanswered ethical/legal questions in this domain
is the question of what rights and affordances will have to be allocated to
robots operating in public spaces.  In a survey on robots in public spaces~\cite{public_spaces_survey}, the
authors highlight seven factors to evaluate such a rollout: safety, privacy and
ethics, productivity, esthetics, co-creation, equitable access, and systemic
innovation, which are closely aligned with the points we sought to highlight in
our teaching module.  In an article on regulation of robots in public spaces~\cite{public_spaces} the author makes the point that
depending on expectations of the function of public spaces can produce widely
different conclusions about what robotic systems might be appropriate to deploy
and how they should be governed.  One very concrete example from the last few
years is that many cities in America have seen significant public pushback to
the rollouts of shared electric scooter infrastructure by private companies.



\paragraph{Relation To Previous Work} In this paper, we contribute a socio-ethical module on responsible robotics -- we draw inspiration from previous ethical CS role-playing modules listed above. Our module goes beyond ethical ML and AI topics to include issues which directly related to the field deployment of autonomous agents. Finally, we directly connect the discussion of socio-ethical factors to technical concepts developed earlier in the course.

\input{sec/overview_course_module}

\input{sec/in_class_role_playing}
\input{sec/outcomes}

\section{Conclusion}
In this paper, we presented a Responsible Robotics teaching module intended for inclusion in upper-division and graduate-level robotics courses.
Evaluations in the form of anonymous surveys indicate that the module was both effective at conveying the importance of the topics discussed, and that the students appreciated the format.
Building on this work, we believe there are several directions for future curricular development --- a longer-running module might benefit from a combination of reflection following the group discussions; providing scenarios for role-play in advance might enable students to form individual opinions before group discussions; and perhaps incorporating some of the technical conflict in the technical assignments might further enhance the learning experience.
We remain convinced that technical courses can benefit from inclusion of discussions on socio-ethical factors, and that serious discussion of complex socio-ethical factors of robotics requires strong technical understanding of the relevant algorithms and hardware.

\section*{Acknowledgements}
This work was supported in part by NSF under grant CAREER-2046955. The views and
conclusions contained in this document are those of the authors only. The
authors thank the anonymous reviewers for their helpful comments. We also thank
the students for their participation and their feedback via the anonymous survey.

\bibliography{aaai22}

\end{document}

%% file: sec/overview_course_module.tex
\section{The Responsible Robotics Course Module}
\seclabel{module}

A recent study by the Knight Foundation found that well meaning mobile delivery robot developers might be designing for sidewalk infrastructure which currently does not reliably exist in cities which they are targeting for deployment~\cite{knight_found}.

A company in such a position would need to make some tough choices about how to
proceed.  For instance, they could go ahead with their plans unchanged, but might
face significant problems with field reliability.  They could try to lobby the
city to upgrade the infrastructure, but this could be a very time-consuming
process, requiring a high degree of community buy-in.  They could choose to
deploy in other places with more favorable operating environments, but such
a decision might lead to disparate impact, and moving would also result in lost
time in valuable markets.

This is the sort of scenario that we hope to prepare students for in this module.  By considering robotic deployments within a broader social context, we set out to prepare students to reason about issues of:  safety, local community sentiment, disparate impact, and labor relations, as they pertain to robotic deployments.  Concretely, we paid particular attention to questions of:
\begin{itemize}
    \item How technical choices map to issues of human safety and system reliability in unstructured environments;
    \item How system design features affect user trust;
    \item How autonomous infrastructure might reinforce existing
    inequalities; and
    \item How labor markets may be impacted by automation.
\end{itemize}

 \subsection{Module Design}

 We chose to focus not on the contradictory constraints governing an individual autonomous agent, but on the competing interests of the human \emph{Business Managers}, \emph{Technology Developers}, \emph{Operators}, and \emph{Users} of real world systems.  More details regarding how these roles are defined can be found in Table~\ref{tab:user_roles}.

Fundamentally our systems are designed by and primarily for other humans, and
this systems-level framing is grounded in how robots are deployed in the real
world today. We hope that exposing students to the inherent conflicts within
such a framework will prepare them to make better choices when they are tasked
with overseeing their own autonomous systems roll-outs.

\begin{table*}[tb]
  \centering
   \begin{tabular}{ |p{2cm}||p{6.5cm}|p{6.5cm}|  }
 \hline
 {\bf Stakeholders} & \emph{Business Management/Strategy}  & \emph{Technology Developers (Engineering/Design)} \\
 \hline
 {\bf Role }   &  Identify the business strategy, monetization modality, company structure and pay scale, market demand, defines recruitment strategy.  & Outline design decisions about the areas of deployment, the sensor, actuator, compute, and algorithm suite for the robots, interaction of the robots with operators and users, safety, security, privacy, and rollout strategy. \\
  \hline
 {\bf Stakeholders} & \emph{Operations} & \emph{User} \\
 \hline
 {\bf Role }   &  Identify the day to day operations strategy for the company, the roles of different human operators, order fulfilment, customer interaction, and failsafe recovery. & Identify who would benefit from the service, their expectations, how much users would be willing to pay for the service, interactions with the system, and restrictions and constraints from bystanders and third parties. \\
 \hline
\end{tabular}
  \caption{Four Roles of Embodied Systems Deployment: We begin our in-class role playing exercise by breaking students up into four groups, each assigned to a stakeholder role.  In the table we summarize the task given to each group.  These stakeholder roles are quite general and thus this exercise could be tailored to many hypothetical systems.}
  \label{tab:user_roles}
\end{table*}




\subsection{Curricular Objectives}
As noted above, any infrastructure project which achieves adoption at scale is necessarily shaped by a number of fairly unforgiving \emph{reality constraints}.  Our principal goal in this module was to encourage students to think about the long term impact of design decisions, subject to such constraints, i.e. to consider potential problems that might arise in systems which do manage to ``beat the odds'' and achieve high levels of adoption.

By guiding students through a series of scenarios we arrive on the last day of the module prepared to consider questions of policy design.  To this end, we discuss questions of how infrastructure level choices of technologists can lead to systems which perpetuate or increase inequality due to rollouts prioritized purely by market forces, or how at scale centralized services can impede individual freedoms by impinging on privacy or forbidding device repair in the terms of service, often in the name of security or ease of use.

\subsection{Structure of Activities}
The core of our teaching module is centered around three in-class role-playing exercises.  One week before these were scheduled, we assigned a written homework exercise.  It contained broad questions regarding potential robotics applications, their technical requirements, social acceptance, and disparate impact.
This module also took place approximately half-way through the semester, so students went into the in-class module having spent some time considering both broader social questions as well as more technical issues relating to robust algorithm design.

\subsection{Reading and Writing Warm-up Assignment}
    To start off, we assigned a list of \emph{suggested reading} including media articles covering deployed robotics applications, essays on the bounds, limitations, and capabilities of AI and robotics, and technical descriptions of deployed systems\footnote{A draft of our course materials can be found together with the source of this publication, or at the links in the footnotes: \url{https://www.cs.utexas.edu/~joydeepb/responsible-robotics-handout.pdf}}. The assignment was released a week before the in-class activities, and was due before the first in-class discussion. We also assigned a number of writing prompts intended to get students thinking about \emph{Applications}, \emph{Technical Requirements}, \emph{Social Acceptance}, and \emph{Disparate Impact} of robotic infrastructure.

    The expected outcome of this assignment was to encourage students to spend a bit of time surveying the landscape of applied robotics.  Our hope was that this assignment would get students to begin thinking about evaluating potential robotics applications along the metrics of technical merit/feasibility as well as societal impact, by encouraging them to identify applications that they personally found compelling or off-putting.  In some sense the purpose of this exercise was to create space for students to grow by exercising their empathy and \emph{ethical intuition}~\cite{Huemer2005-HUEEI}, without getting too far afield in philosophical texts and ethics frameworks during a highly technical robotics course.

%% file: sec/in_class_role_playing.tex
\section{In-Class Role-Playing Activities}
\seclabel{activities}

    The core structure of the Responsible Robotics module was aimed at getting students to role-play corporate life, first by having students embody a role, and next by doing so in opposition to the competing interests of actors in other roles.  On the third day we come together as a class and have each team consider the same set of plausible social changes which these technologies have the potential to precipitate, colored by the role they had played on the previous two days.
    \subsection*{Example Scenario: Operation Medibot}
    In our class we based everything on considering a hypothetical startup idea in the space of pharmaceuticals delivery.  In particular, we asked students to consider the premise of leveraging recent advances in robotics to start a modern delivery service called {\bf MediBots}:
    \begin{quote}
        \emph{MediBots} promises to deliver medications straight to people’s homes while offering significant savings over brick and mortar stores like CVS and RiteAid.
MediBots is planning to deploy a fleet of semi-autonomous vehicles to drive packages up to buildings in neighborhoods, which are then delivered to people's doorsteps and mailboxes by quadrotor drones.
    \end{quote}

    We intentionally chose a fictitious example scenario most likely to be fraught with a wide range of ethical issues while still being relatable as a realistic startup, given the recent spate of last-mile delivery startups.

    \subsection{Day 1: The Startup Crunch}
    The curricular objective of the first day was to internalize stakeholder
    roles and technical design criteria\footnote{\url{https://www.cs.utexas.edu/~joydeepb/responsible-robotics-day1.pdf}}. To do this, students met in groups to discuss
    the questions below, and then reconvened to share ideas for the second half
    of the class.
   Importantly, these discussions were grounded in the students experience of implementing an obstacle avoidance planner and a mobile robot localization algorithm on actual scale 1/10 cars in previous weeks.

    \subsubsection{Group Discussion Questions}
    \begin{itemize}
        \item  \emph{Business Management / Strategy}: How will the business idea be sold to investors? How will it be financially viable and sustainable?  What are potential obstacles that such a business would have to overcome in order to scale?  How could such a business differentiate itself in the marketplace in the face of larger players with more resources moving into this vertical?
        \item  \emph{Engineering and Execution}: Identify safety, privacy issues and concerns with this idea. How would you test, train, and deploy robots while maintaining privacy? Recall that remote operators might need to see live sensor feeds to help robots.  How would training data be acquired, and what would it be used for?  How would data be secured?
        \item  \emph{Operations Group}: How will orders come in, how will they be dispatched, how will the interface with pharmacies and patients respect privacy?  Where might human labor be needed in order to augment automation?
        \item \emph{Users Group}: Who will find this product useful? What expectations will they have for it?  What legal/moral responsibilities would MediBots have in handling this sensitive user data?  How will this effect community members who are not in the target market for the service?
    \end{itemize}

    \subsection{Day 2: Competing Reality Constraints}
    The curricular objective of the second day was to learn to recognize the
    tension between competing interests and constraints inherent to real-world deployments\footnote{\url{https://www.cs.utexas.edu/~joydeepb/responsible-robotics-day2.pdf}}.
    We asked students to consider solutions to the following set of competing interests from different stakeholders.  All students answered the same set of questions, but group responses were colored by each teams discussions from day 1.

        \subsubsection{Scenario 1: Financial Constraints. }
Consider the conflicting consequences from cutting various corners and costs in your autonomous delivery product.

1.1 The engineering team must cut the price of its delivery drones in half.  What consequences might this have on operating the robot fleet?

1.2 The operations team is now tasked with processing twice as many manual interventions as they were planning for, but there is no budget to hire additional ops staff, what are potential consequences of this?  How might the engineering team roll out features to address these problems?

1.3 Management has raised additional capital and is attempting to deploy delivery vehicles nationwide in an attempt to establish itself in new markets ahead of the competition, but has no staff on the ground in these markets to maintain its autonomous vehicles.  Where might MediBots look to hire staff for its nationwide distributed fleet maintenance team?

1.4 The engineering team has been tasked with removing the expensive LIDAR sensor from the autonomous vehicle and replacing it with stereo RGB cameras, and the Ops team has been tasked with increasing the uptime of autonomous deliveries by 20\% per team member.  What will have to give?

\subsubsection{Scenario 2: Competing Incentives.  }
Users rave about the convenience of MediBots, but unbeknownst to them, the management team has entered into a data sharing agreement with several large advertising conglomerates in an effort to increase quarterly earnings ahead of their next fundraising round.

2.1 What are some potential long term implications of this business decision?

2.2 What are the concerns with this arrangement? Is there any way this can be done in an acceptable manner?

\subsubsection{Scenario 3: Community Pushback}
The dream that communities would welcome MediBots with open arms has not played out as expected, and there are several sources of tension between MediBots and the communities it serves.

3.1 MediBots has noticed that in certain neighborhoods there are high rates of package theft, what policies should it enact to improve yields?  What are some potential impacts of these policies on the residents of these neighborhoods?  Do you feel your proposed policy is equitable?

3.2 MediBots has been incurring heavy losses of its autonomous vehicles, where people seem to be intentionally damaging robots/drones while they are in the field.  What interventions should the company consider to lower costs?  How will your proposed intervention affect the quality of life in neighborhoods where it is deployed?

3.3 The vibrant biking community of Austin, Texas is up in arms about the MediBots ground robots taking up entire bike lanes (where available), and ambling along at 5mph. Things have come to a head, and the biking community is threatening to block passage of the MediBots across town. Luckily a community open house is scheduled between MediBots and Austin bikers to discuss the situation.
Company management: What will you bring to the meeting to assuage the public?
Users group: What changes, assurances, and information would you like the MediBots execs to provide?

\subsection{Day 3: The Price of Success}

On Day 3 of the module, we concluded with a discussion of broader impacts and
trends, considering how the introduction of a service into a community may alter
its social fabric\footnote{\url{https://www.cs.utexas.edu/~joydeepb/responsible-robotics-day3.pdf}}.  Concretely, there is widespread concern about robotics
disrupting the workforce and potentially negatively impacting
lives and livelihoods in the short term.  We feel that it is a productive
use of class time to ask robotics students to consider the likelihood of such
adverse effects, and the ethical ramifications
of this.  Is this a real concern, or a misplaced societal fear? Is this simply the cost of progress, or should some support structures
exist to help lower the social costs of the coming shifts in the labor market?
What can we learn from the social rifts perpetuated by the industrial
revolution, and can we do better in the future?

\subsubsection{Discussion}

We also extrapolate from examples of how simple market forces today can breed
inequality, \eg{} consider the prevalence of \emph{food deserts} in low income
urban areas.  Here, we discuss how autonomous agents are high value targets full
of expensive components like sensors, motors, and batteries.  One might imagine
that the companies or governments controlling autonomous infrastructure could
very naturally limit deployments in areas with higher crime rates.  As
autonomous infrastructure is likely to at least partially displace older modes
of resource distribution, fabrication, and transportation, this has the
potential to negatively impact the lives of consumers left behind.  We raised
the question of who should ultimately be responsible of ensuring that infrastructures are rolled out
in an egalitarian manner --- through self-regulation, market forces, or via
government policy.

%% file: sec/outcomes.tex
\section{Outcomes}
\seclabel{evaluation}
We solicited anonymous survey responses at the conclusion of the course module to assess the sentiment of the students at the conclusion of the activity, and to gauge the effectiveness of the module.
Our survey consisted of a mix multiple choice questions to measure sentiment as well as two short response fields for collecting more free form answers.
The results of anonymous surveys, which we present in this section, have been encouraging, leading us to think about how such a module may be re-used for other robotics courses, and how we might further improve upon it.






\subsection{Anonymous Student Surveys}

In our graduate class, we received 33 survey responses, and in our undergraduate class received 28 responses.

\begin{figure*}[t!]

\centering
\begin{tabular}{c c c}
\begin{subfigure}{0.3\textwidth}
  \includegraphics[width=\textwidth]{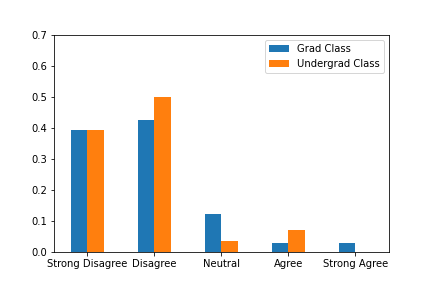}
  \caption{I believe roboticists should just focus on building robots, and leave the issues of how they are used and deployed to others.}
  \label{fig:3}
\end{subfigure} &
\begin{subfigure}{0.3\textwidth}
  \includegraphics[width=\textwidth]{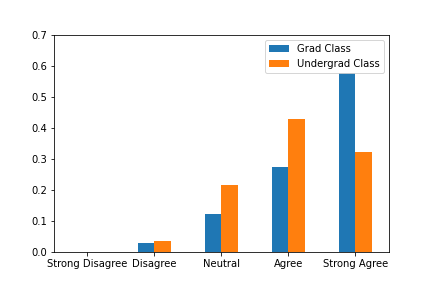}
  \caption{I feel it is my responsibility to raise concerns about how technology is used by a company I work for.}
  \label{fig:6}
\end{subfigure} &
\begin{subfigure}{0.3\textwidth}
  \includegraphics[width=\textwidth]{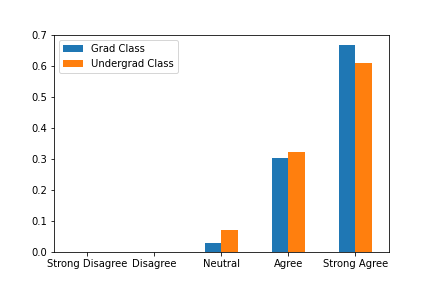}
  \caption{I believe it is important for roboticists to consider the ethical and social implications of the deployment of robots.}
  \label{fig:4}
\end{subfigure}\\

\begin{subfigure}{0.3\textwidth}
   \includegraphics[width=\textwidth]{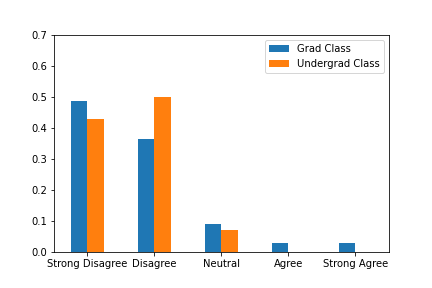}
   \caption{I believe engineers need not concern themselves with how the technology they develop affects society - that is a job for others.}
\end{subfigure} &
\begin{subfigure}{0.3\textwidth}
   \includegraphics[width=\textwidth]{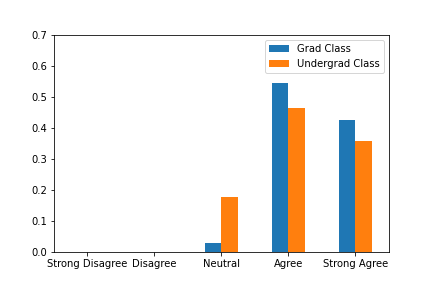}
   \caption{I have discovered social and ethical factors, and/or new angles to previously known factors,  related to deploying real-world robotic systems.}
   \label{fig:2}
\end{subfigure} &
\begin{subfigure}{0.3\textwidth}
  \includegraphics[width=\textwidth]{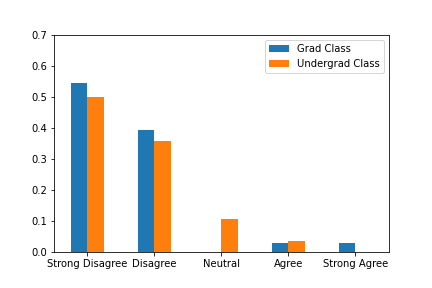}
  \caption{I feel I could have learned more about these issues working on my own than in a group. \\}
  \label{fig:10}
\end{subfigure} \\

\begin{subfigure}{0.3\textwidth}
  \includegraphics[width=\textwidth]{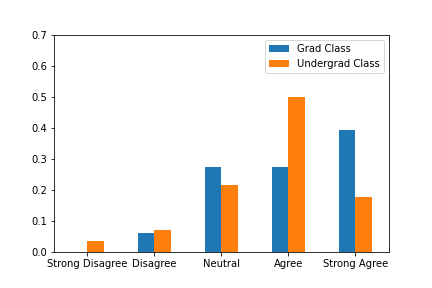}
  \caption{I found the responsible robotics writing assignment useful in exposing me to issues discussed in class.}
  \label{fig:8}
\end{subfigure} &
\begin{subfigure}{0.3\textwidth}
  \includegraphics[width=\textwidth]{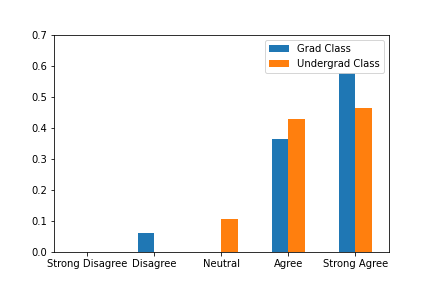}
  \caption{I found the group discussions helpful in forming opinions and discussion the issues raised in class}
  \label{fig:9}
\end{subfigure} &
\begin{subfigure}{0.3\textwidth}
  \includegraphics[width=\textwidth]{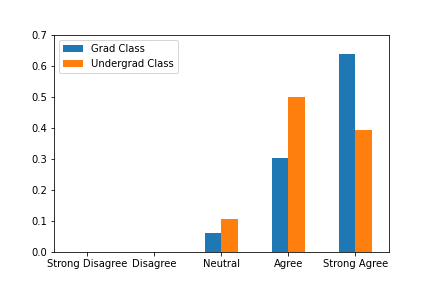}
  \caption{I believe robotics courses should include discussions about the social and ethical implications of robot deployments.}
  \label{fig:13}
\end{subfigure}

\end{tabular}

  \caption{  Summary of survey responses from Responsible Robotics offered to two student cohorts.  See the outcomes section for a more complete discussion of these survey results.  }
  \label{fig:data}
  \vspace*{-0.5cm}
\end{figure*}

The questions we asked in this survey helped us gauge the effectiveness of the Responsible Robotics module along axes of engagement, effectiveness and outcomes.  Specifically, we asked students to complete surveys with a number of agree/disagree multiple choice questions.  The responses we received showed that the class was in consensus about this module being a productive use of time. We include an aggregated view of our survey results in Figure~\ref{fig:data}.


\subsubsection{Student Outcomes}
We found that students almost uniformly came away feeling that roboticists have an imperative to weigh in on the broader social impacts of robotics, and that it is their job to consider the social impacts of technology that they develop.  We unfortunately did not collect data about student sentiment to this point before running the module, but what we can say is that in the exit survey, students responded almost unanimously reported that they:
\begin{itemize}
    \item gained a new appreciation for technical challenges they might encounter in the field
    \item gained a new appreciation for social and ethical factors which might be impacted by technical decision making
    \item came away empowered, feeling that considering these issues is part of a roboticists job.
\end{itemize}

\subsubsection{Effectiveness Of The Module}
The students also came away saying that they enjoyed the inclusion of this module in the course, and that they found its structure to be effective.  In particular, students reported that they:
\begin{itemize}
    \item gained knowledge technically relevant to deploying robots in the field
    \item found the group role-playing setting to be useful in terms of learning more ideas and forming opinions.
    \item they found the assigned warm-up writing assignment to be useful.
    \item felt that modules like this should appear as a standard part of robotics courses.
\end{itemize}

\subsubsection{Student Engagement}
We ran this assignment in a hybrid setting where students in each group were remote (in the graduate class) as well as in a fully in-person setting (for the undergraduate class).  We found on the first day student engagement was highly group dependent, but increased uniformly on subsequent days.

We had student groups write their responses into a shared document.  When surveyed, we found that 85.7\% in the undergrad class and 78.8\% of the graduate students had read through the contributions of the other groups to the in-class exercises. Of the people that read the contributions, most of them found it a productive use of time.

\subsection{Open-Ended Student Responses}

We also included two open answer fields on the survey to solicit free-form responses.  Here we include a collection of representative responses to these questions. 

 \subsubsection{``I believe a responsible roboticist is one who...''}

\begin{itemize}
    \item \emph{``Considers their social impacts when designing and manufacturing robots''}
    \item \emph{``Thinks about the societal and ethical implications of technology before they create it''}
    \item \emph{``Considers the negative externalities of the decisions they make, and actively works to combat them.''}
    \item \emph{``Not only builds robust and efficient robots but also pays attention to the consequences of the development of these robots.''}
\end{itemize}





\subsubsection{Open Answer}

Here we provide a few representative comments from the ``comments / questions / suggestions?'' field on the survey.

\begin{itemize}
    \item \emph{``I did not expect this assignment, but I am glad that it was given.''}
    \item \emph{``This was a great week, I really appreciated this since it's pretty much the first time a CS class has focused on the ethics of the technology rather than just technical implementation details.''}
    \item \emph{``Thanks for doing this discussion! I feel like the ethical ramifications of our work as computer scientists in lots of different areas can be glossed over in classes, but we can have such a big impact on the world without realizing it.''}
    \item \emph{``This was a very nice break from the theory presented normally in lecture. It has helped me a lot to open my eyes to what great responsibility this generation of roboticists have. We are setting the stage for what the world will look like many generations away.''}
\end{itemize}

\subsection{Instructor Reflections}


The writing assignments and in class prompts were carefully designed so that more clear cut questions appeared first, and more ambiguous moral questions appeared later on, to encourage students to more honestly and fully engage with the presented scenarios.

For the in-class discussions, we considered working through multiple role playing scenarios.  We feel that the choice of sticking to a specific scenario that played out in detail allowed students to engage more deeply with the material. We also deliberately chose an example that would be plausible, but which would have the potential to develop into large-scale societal issues raised on day 3. Framing the setting as a startup was quite appealing to the students and allowed them to engage deeply right from the outset. The chosen example was also carefully crafted to have the potential of positive impact (medicine delivery), but still come with deep socio-ethical issues during deployments. We conjecture that an adversarial setting (\eg{} battle robots) would have been immediately off-putting and would have inhibited deeper discussions. On Day 1, we noticed that all groups were very engaged in the idea and took their roles seriously to lay out detailed plans.

 We believe that it was particularly valuable for students to consider how those who stand to benefit the \emph{least} from the roll-out might be affected anyway by it. Future iterations might consider experimenting with introducing such a fifth stakeholder group in the interest of promoting harm reduction as a critical design principle for public infrastructure projects.


An in-class discussion on Day 3 on how to raise objections to questionable practices in your company was very well-received. There were several good ideas, including escalations of reporting, finding peer groups of similar-minded employees, advocating for third-party company ombudspersons, looking out for companies that actively welcome opinions, and joining small companies where employees have more say.

The in-class activities had more questions than the students could thoroughly discuss in the allotted 20 minutes. In general, more time for discussion would have been good for the first and second days.
The third day had three sections: the group discussions, an individual questionnaire, and a whole class discussion on actionable strategies.

In practice, perhaps the most useful take away a student could come away from this exercise feeling, is emboldened to resist implementing something that they themselves find unethical, and having a sharper understanding of their own morality.

Having the instructors sit in with the groups was extremely helpful -- we were able to coax quiet students into participating, and were able to highlight angles that the students might have missed, or the general high-level insight behind the questions, that is documented in this publication.